\documentclass[11pt]{article}

\usepackage[preprint]{acl}

\usepackage{multirow} 
\usepackage{booktabs} 
\usepackage{listings} 

\usepackage{xcolor} 

\usepackage{times}
\usepackage{soul}
\usepackage{url}
\usepackage[utf8]{inputenc}
\usepackage{graphicx}
\usepackage{amsmath}
\usepackage{amsthm}
\usepackage{booktabs}
\usepackage{algorithm}

\usepackage{amssymb}
\usepackage{algpseudocode}
\usepackage{bm} 

\usepackage[T1]{fontenc}

\usepackage[utf8]{inputenc}

\usepackage{microtype}

\usepackage{inconsolata}

\usepackage{graphicx}

\usepackage[switch]{lineno}

\title{Progressive Content Refinement with Decaying Reward Joint LinUCB}

\author{
  Shion Ishikawa \\
  Rakuten Group, Inc. \\
  Tokyo, Japan 
  \And
  Pablo Loyola \\
  Rakuten Group, Inc. \\
  Tokyo, Japan 
  \And
  Young-joo Chung \\
  Rakuten Group, Inc. \\
  San Mateo, USA 
  \And
  Yun Ching Liu \\
  Rakuten Group, Inc. \\
  Tokyo, Japan 
}

\begin{document}

\maketitle

\begin{abstract}
Iterative refinement has significantly enhanced Large Language Model (LLM) performance; however, existing methods—ranging from feedback-based Self-Refine to traditional bandit approaches—often rely on static options or overlook the ``saturation effect''. This neglect leads to over-exploitation, where the continuous use of identical prompts or arms results in diminishing rewards over time.

To address this challenge, we propose a novel contextual bandit algorithm that explicitly incorporates reward decay modeling. Utilizing an Expectation-Maximization (EM) algorithm, our method simultaneously estimates both arm-specific and decay parameters. Furthermore, by embedding prompts as arms, we facilitate the joint learning of arm values, distinguishing our approach from the traditional disjoint Linear Upper Confidence Bound (LinUCB) framework.

Experimental results on Sentiment Reversal and GSM8K benchmarks demonstrate that our method achieves significant performance gains over strong baselines. Finally, our ablation study confirms that the integration of reward decay modeling within the bandit framework is crucial for mitigating over-exploitation and optimizing the iterative refinement process.

\end{abstract}

\section{Introduction}

In recent years, Large Language Models (LLMs) have revolutionized the field of natural language processing, demonstrating remarkable performance across a wide array of tasks such as text generation, summarization, translation, and complex reasoning \cite{llm,10.5555/3722577.3722647}. To further unlock the full potential of these models and obtain higher-quality, more reliable outputs, iterative refinement techniques have become increasingly crucial \cite{self-refine,iterative_clinical}. These methods aim to systematically enhance LLM-generated content by applying successive improvements based on various feedback mechanisms.

While initial iterative refinement approaches like Self-Refine \cite{self-refine} have shown promise by allowing LLMs to critique and improve their own outputs, they often rely on pre-defined or heuristically chosen refinement prompts. This static selection of prompts can lead to suboptimal performance, as the effectiveness of a refinement strategy can vary significantly across different contexts and over time. More sophisticated methods have begun to incorporate multi-armed bandit (MAB) frameworks. These frameworks treat different refinement strategies (e.g., prompt design strategies, contents to refine) as "arms" to be pulled \cite{bandit-prompt-strategy,rex}. These bandit-based approaches effectively balance the exploration of novel refinement strategies with the exploitation of known effective ones, optimizing the overall refinement process.

However, a critical limitation in existing iterative refinement paradigms, particularly prevalent in bandit-based methods, is the implicit assumption of stationary reward distributions for each refinement strategy (arm). In reality, repeatedly applying the same refinement prompt or exploiting a single arm can lead to a "saturation effect." For instance, an LLM might quickly exhaust the utility of a specific grammatical correction prompt after a few iterations, or a mathematical reasoning prompt might become redundant once a certain level of accuracy is achieved. Continuing to apply such a prompt results in diminishing returns, or even negative impacts, akin to over-editing. Empirical evidence for this decay phenomenon is presented in Figure 4 of \cite{self-refine}. This phenomenon, termed "reward decay" by prior work like Rotting Bandit \cite{rottingbandit}, leads to inefficient over-exploitation of diminishingly effective strategies, hindering overall refinement efficiency. Traditional bandit algorithms, designed for stationary environments, fail to adequately capture and adapt to this dynamic decay in reward.

While the concept of decaying rewards has been explored by Rotting Bandit, this approach suffers from inefficiencies due to their initial exploration strategies, such as relying on round-robin sampling to estimate decay rates. 

To address these fundamental challenges, we propose a novel bandit algorithm that explicitly models and adapts to the reward decay phenomenon in LLM iterative refinement. Furthermore, our approach abstracts arms as embeddings, treating these as the context for each arm. Traditional LinUCB-like approaches \cite{linucb} often have arm features that are disjoint with each other (e.g., assuming distinct arm models for how arm context affects click-through rates in sports and political news recommendations). In contrast, our method assumes a unified underlying model regarding how prompt embeddings influence rewards. This means that when a new prompt (arm) is generated, its embedding representation allows for a more informed initial estimation of its parameters, effectively mitigating the LLM-specific ``cold start'' problem often encountered with new prompts.

Finally, by introducing an Expectation-Maximization (EM) algorithm \cite{emalogrithm}, we achieve simultaneous learning of both decay and contextual parameters, thereby circumventing the inefficient exploration phase associated with round-robin sampling in methods like Rotting Bandit.

These techniques allow our system to dynamically balance exploration of potentially effective but less-used prompts with the exploitation of currently high-performing prompts, while actively accounting for their diminishing utility over time.

Our technical contributions are threefold:

\begin{itemize}

\item We propose a novel bandit algorithm DR-LinUCB \footnote{ Source code attached for review. GitHub repository will be available at https://github.com/\{anonymous\_org\} upon acceptance.} that explicitly incorporates a reward decay model for each arm, enabling adaptive exploration-exploitation in dynamic LLM refinement environments.

\item We introduce a method to embed refinement prompts as contextual features for our bandit arms, enabling the joint learning of arm values. Unlike traditional disjoint LinUCB, our method assumes a unified arm model for each LLM task, which allows for a more informed initial estimation of parameters for new prompts, thereby mitigating the LLM-specific "cold start" problem.

\item We leverage an Expectation-Maximization (EM) algorithm for the simultaneous learning of both arm contextual parameters and their associated decay parameters from observed rewards. This approach avoids inefficient initial exploration strategies such as round-robin sampling.

\end{itemize}

\begin{figure}[htbp] 
    \centering 
     \hspace*{-0.2cm}
    \includegraphics[width=0.49\textwidth]{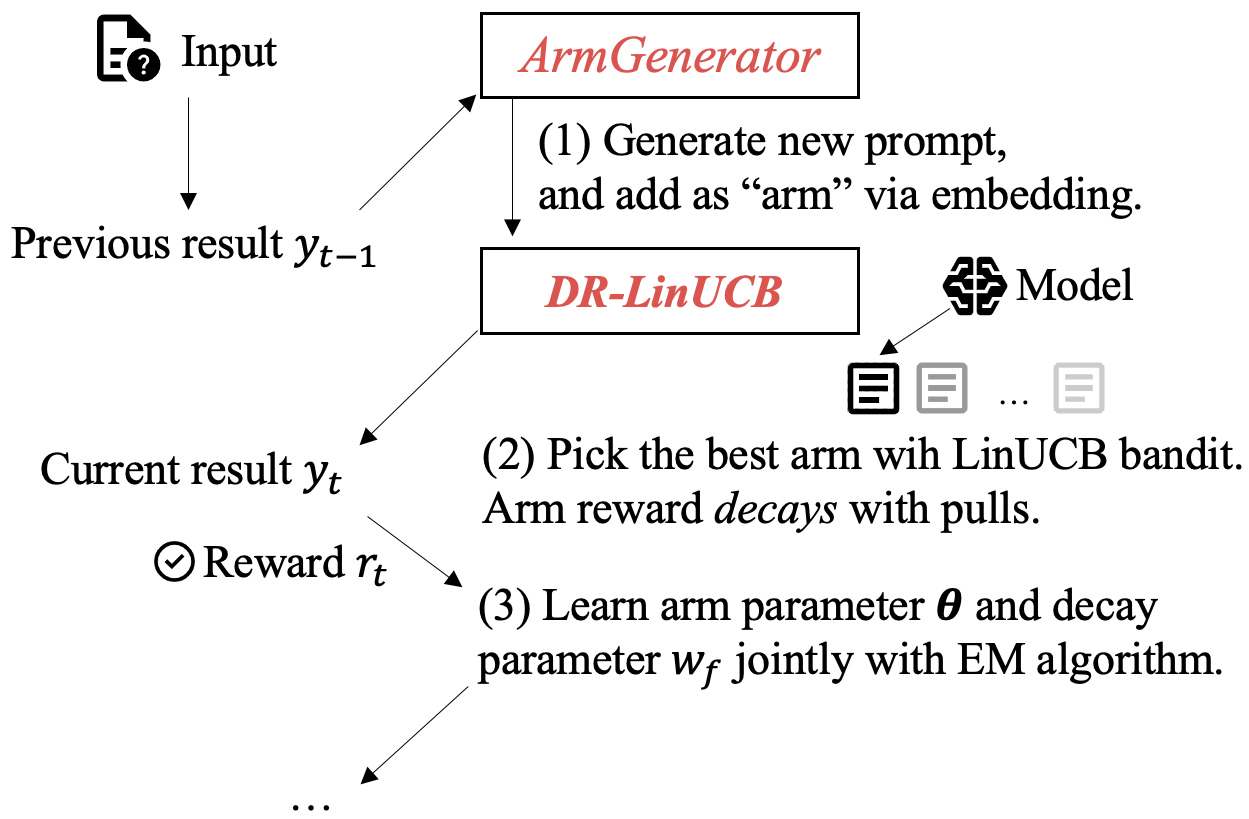}
    
    \caption{DR-LinUCB's Progressive Content Refinement. This flowchart details its core components: reward decay modeling, UCB-based arm selection and joint EM-based parameter learning.}
    \label{fig:score_history_combined} 
\end{figure}

\section{Related works}
\subsection{Iterative Refinement for Large Language Models}
The ability of LLMs to perform iterative self-correction has emerged as a key technique for enhancing generation quality.
SELF-REFINE \cite{self-refine} introduced LLMs generating, critiquing, and refining their own outputs using natural language feedback, significantly improving performance across tasks. This concept extended to specific domains: Chen et al. (2024) applied it to machine translation for human-preferred fluency \cite{chen-etal-2024-iterative} , and Hein et al. (2025) developed high-accuracy clinical information extraction pipelines, emphasizing task definition and human-in-the-loop refinement \cite{iterative_clinical} . Despite these advances, a key limitation of existing methods is their limited prompt diversity, which our work mitigates through bandit explore-exploitation.

\subsection{Bandit Algorithms for Large Language Model}
In multi-step tasks, LLMs need to balance using proven prompts (exploitation) with trying new ones (exploration). Bandit algorithms help by intelligently guiding this choice, preventing over-reliance on old prompts and efficiently discovering better ones.

In code refinement, the REx (REfine, Explore, Exploit) algorithm \cite{rex} frames iterative LLM-based code repair as an arm-acquiring bandit problem. Each generated code is an arm, and refining it yields a reward (passing tests). REx uses Thompson Sampling with heuristic-informed priors to balance exploring new code and exploiting promising ones. It significantly improved problem-solving and reduced LLM calls across diverse coding tasks.

For prompt optimization, OPTS (Optimizing Prompts with sTrategy Selection) \cite{bandit-prompt-strategy} explicitly selects prompt design strategies using bandit algorithms. Traditional methods often rely on LLMs implicitly choosing strategies, which can be suboptimal. OPTS(TS), its most effective variant, treats each strategy as an arm, learning its efficacy via Thompson Sampling \cite{thompsonsampling}. This approach improved prompt performance by up to 50\% by making strategy selection adaptive and data-driven.

Unlike REx and OPTS, which respectively refine content candidates with a fixed prompt and optimize a single prompt, our approach dynamically selects generated prompts. This enables prompt selection to be adapted based on the current output state, allowing for contextual content refinement.

\subsection{Bandit Algorithms with Decaying Rewards}
The "Rotting Bandits" framework \cite{rottingbandit} addresses Multi-Armed Bandit (MAB) problems where an arm's expected reward decays with its pull count, departing from classical stationary assumptions. They proposed non-parametric methods like Sliding-Window Average and parametric approaches such as Closest To Origin for this setting. However, their reliance on initial round-robin exploration for decay rate estimation can be inefficient. 

Our work enhances this by introducing an Expectation-Maximization (EM) algorithm, enabling simultaneous learning of decay and contextual parameters, thus circumventing the inefficient initial exploration inherent in traditional Rotting Bandit approaches.

\subsection{Action embeddings}
Large action spaces demand efficiency and scalability. action embeddings schemes address this by simplifying action representations. Action embedding is vital for Off-Policy Evaluation (OPE). Marginalized IPS (MIPS) \cite{saito2022offpolicyevaluationlargeaction} uses embeddings to reduce OPE variance in large action spaces, enhancing generalizability. Furthermore, action embeddings are useful in position bias estimation \cite{poisionbias_item_embedding}.

Our work applies action embeddings for efficiency in large LLM prompt spaces. Further, using action embeddings within an arm-acquiring bandit framework, we enable adaptation to new ``arms'' (e.g., novel prompt strategies), facilitating robust LLM optimization.


\section{Problem Setting}
We consider an arm-acquiring contextual bandit problem with decaying rewards, where an agent's objective is to maximize cumulative reward over $T_{\text{max}}$ timesteps. At each timestep $t$, an agent selects an arm $i(t)$ from the available set $\mathcal{A}_t$ according to a policy $\pi$. Each arm $i \in \mathcal{A}_t$ is characterized by a $d$-dimensional embedding $\mathbf{e}_i \in \mathbb{R}^d$. We assume these arm embeddings lie within a compact subset of $\mathbb{R}^d$, for example, a unit sphere or a hypercube. The set of available arms $\mathcal{A}_t$ can dynamically expand over time ($\mathcal{A}_{t+1} \supseteq \mathcal{A}_t$) with arm acquisition. Upon selection, the agent observes a real-valued reward $r_t \in [0, R_{\max}]$ for some $R_{\max} > 0$.

The observed reward $r$ for arm $i$, after being pulled $N_i$ times, follows an exponentially decaying model:
\begin{equation} r = \tilde{\mu}_i \cdot \exp(-N_i \cdot f_i) \end{equation}
where $\tilde{\mu}_i = \mathbf{e}_i^\top \bm{\theta}$ is the initial expected reward (linearly modeled by a global parameter $\bm{\theta} \in \mathbb{R}^d$), and $f_i = \max(0, \mathbf{e}_i^\top \mathbf{w}_f)$ is the decay rate (determined by a global decay parameter $\mathbf{w}_f \in \mathbb{R}^d$).

The agent's ultimate objective is to find a policy $\pi$ that maximizes the cumulative reward $\sum_{t=1}^{T_{\text{max}}} r_t$, equivalent to minimizing the cumulative regret $R(T_{\text{max}})$:
\begin{equation} R(T_{\text{max}}) = \sum_{t=1}^{T_{\text{max}}} \left( r_{t}^{*} - r_t \right) \end{equation}
Here, $r_{t}^{*}$ denotes the reward from an oracle selecting the optimal arm at timestep $t$ considering its current decay.


\section{Decaying Reward Joint LinUCB (DR-LinUCB)}

DR-LinUCB is the core arm selection and parameter learning algorithm. It integrates a reward decay model and an EM algorithm into a joint linear UCB framework, where parameters are shared or learned across arms through their embeddings. Pseudo algorithm of DR-LinUCB is given by Algorithm 1.

DR-LinUCB consists of several key elements, detailed below sections.

\subsubsection{UCB Selection with Reward Decay}

At each timestep $t$, DR-LinUCB selects an arm $i(t) \in \mathcal{A}_t$ from the available arms that maximizes the following criterion:

\begin{equation} i(t) = \operatorname{argmax}_{i \in [K_t]} \left( \mathbf{e}_i^\top \hat{\bm{\theta}}  + C  \sqrt{\mathbf{e}_i^\top \mathbf{A}^{-1} \mathbf{e}_i}\right)\cdot d_i \end{equation}
where $d_i = \exp\left(-N_i \max(0, \mathbf{e}_i^\top \mathbf{w}_{f})\right)$. 

The first term, $\mathbf{e}_i^\top \hat{\bm{\theta}}$, represents the estimated undecayed expected reward for arm $i$, based on its embedding $\mathbf{e}_i$ and the globally learned parameter $\hat{\bm{\theta}}$. The second term, $C \sqrt{\mathbf{e}_i^\top \mathbf{A}^{-1} \mathbf{e}_i}$, is the exploration term (UCB term) that accounts for uncertainty, where $C$ is a hyperparameter controlling the degree of exploration. Finally, the exponential term, $\exp\left(-N_i \max(0, \mathbf{e}_i^\top \mathbf{w}_{f})\right)$, models the decay of reward based on the number of times arm $i$ has been pulled, $N_i$, and its specific decay rate $\mathbf{e}_i^\top \mathbf{w}_{f}$. The $\max(0, \cdot)$ operation ensures the decay factor is non-negative. This comprehensive criterion considers the saturation effect from excessive exploitation, thereby encouraging the algorithm to explore new arms when existing ones show diminishing returns.

\subsubsection{Expectation-Maximization (EM) Algorithm for Parameter Learning}

DR-LinUCB employs an EM algorithm to simultaneously learn both the global reward model parameters $\hat{\bm{\theta}}$ and the global decay model parameters $\mathbf{w}_f$. This approach is particularly effective for estimating the hidden variable, the undecayed reward $\tilde{\mu}_{i(t)}$, from the observed decayed reward $r_t$.

\begin{algorithm}[H]

\caption{Decaying Reward Joint LinUCB (DR-LinUCB)}

\label{alg:dr_jointlinucb}

\begin{algorithmic}[1]

\Require $\mathcal{A}_0, h_0, T_{\text{max}}, C, \lambda, \lambda_f, \epsilon_{\text{log}}, \epsilon_{\text{decay}}$ 

\State \textbf{Initialize:} $N_i = 0$ for $i \in \mathcal{A}_{0}$
\State $\mathbf{A} = \lambda \mathbf{I}_d$, $\mathbf{b} = \mathbf{0}_d$, $\hat{\bm{\theta}} = \mathbf{0}_d$ 

\State $\mathbf{A}_{f} = \lambda_f \mathbf{I}_d$, $\mathbf{b}_{f} = \mathbf{0}_d$, $\mathbf{w}_{f} = \mathbf{0}_d$ 

\For{$t = 1, 2, \dots, T_{\text{max}}$}
    \State $\mathcal{A}_t \gets \mathcal{A}_{t-1} \cup \text{ArmGenerator}(\mathcal{A}_{t-1}, h_{t-1})$ 
    \State $\bm{E} \gets \{\text{Embed}(a) \mid a \in \mathcal{A}_t\}$ 
    \State $d_i = \exp\left(-N_i \max(0, \mathbf{e}_i^\top \mathbf{w}_{f})\right)$
    \State $i(t) \gets \operatorname{argmax}_{i \in \mathcal{A}_t} ( \mathbf{e}_i^\top \hat{\bm{\theta}}  + C  \sqrt{\mathbf{e}_i^\top \mathbf{A}^{-1} \mathbf{e}_i}) \cdot d_i$ 

    \State Observe reward $r_t$ for arm $i(t)$

    \While{ $\hat{\bm{\theta}}$, $\mathbf{w}_{f}$ not converged} 

        \State $\tilde{\mu}_{i(t)} \gets r_t / d_{i(t)}$ 

        \State $\hat{\bm{\theta}} \gets (\mathbf{A} + \mathbf{e}_{i(t)}\mathbf{e}_{i(t)}^\top)^{-1} (\mathbf{b} + \tilde{\mu}_{i(t)} \mathbf{e}_{i(t)})$  

        \If{$r_t > 0$ and $\tilde{\mu}_{i(t)} > 0$ and $r_t \neq \tilde{\mu}_{i(t)}$}

            \State $y \gets \log(\tilde{\mu}_{i(t)} + \epsilon_{\text{log}}) - \log(r_t + \epsilon_{\text{log}})$

            \State $\mathbf{x} \gets N_{i(t)} \mathbf{e}_{i(t)}$

            \State $\mathbf{w}_{f} \gets (\mathbf{A}_{f} + \mathbf{x}(\mathbf{x})^\top)^{-1} (\mathbf{b}_{f} + y  \mathbf{x})$   

        \EndIf

    \EndWhile

    \State $\mathbf{A} \gets \mathbf{A} + \mathbf{e}_{i(t)}\mathbf{e}_{i(t)}^\top$

    \State $\mathbf{b} \gets \mathbf{b} + \tilde{\mu}_{i(t)}  \mathbf{e}_{i(t)}$

    \State $\mathbf{A}_{f} \gets \mathbf{A}_{f} + \mathbf{x}(\mathbf{x})^\top$

    \State $\mathbf{b}_{f} \gets \mathbf{b}_{f} + y  \mathbf{x}$

    \State $h_t \gets h_{t-1} \cup \{i(t), r_t\}$

    \State $N_{i(t)} \gets N_{i(t)} + 1$

\EndFor

\end{algorithmic}

\end{algorithm}

The E-Step (Line 11) uses the current global decay model parameters $\mathbf{w}_{f}$ to \emph{undecay} the observed reward $r_t$, thereby estimating the undecayed reward $\tilde{\mu}_{i(t)}$. The calculation is performed as $ \tilde{\mu}_{i(t)} \gets r_t / \exp\left(-N_{i(t)} \cdot \max(0, \mathbf{e}_{i(t)}^\top \mathbf{w}_{f})\right) $. A small constant $\epsilon_{\text{decay}}$ is introduced for numerical stability, treating the decay rate as 1 if it's very small.

The M-Step (Lines 12-16) then updates the global reward model parameters $\hat{\bm{\theta}}$ and the global decay model parameters $\mathbf{w}_{f}$ using the $\tilde{\mu}_{i(t)}$ estimated in the E-Step. Specifically, $\hat{\bm{\theta}}$ is updated using standard linear ridge regression, with the undecayed reward $\tilde{\mu}_{i(t)}$ as the target: $ \hat{\bm{\theta}} \gets (\mathbf{A} + \mathbf{e}_{i(t)}\mathbf{e}_{i(t)}^\top)^{-1} (\mathbf{b} + \tilde{\mu}_{i(t)} \cdot \mathbf{e}_{i(t)}) $. The decay model parameters $\mathbf{w}_{f}$ are updated by transforming the exponential decay into a linear regression problem, a process detailed in the subsequent subsection.

\subsection*{Transformation to a Linear Regression Problem for $\mathbf{w}_f$ Estimation}

To estimate the global decay parameter $\mathbf{w}_f$, the exponential decay model $r_t = \tilde{\mu}_{i(t)} \cdot \exp\left(-N_{i(t)} \cdot (\mathbf{e}_{i(t)}^\top \mathbf{w}_f)\right)$ is transformed into a linear regression problem. Taking the natural logarithm of both sides yields:

\begin{align}
    \log(r_t) = \log(\tilde{\mu}_{i(t)}) - N_{i(t)} \cdot (\mathbf{e}_{i(t)}^\top \mathbf{w}_f)  \nonumber \\
    \log(\tilde{\mu}_{i(t)}) - \log(r_t) = N_{i(t)} \cdot (\mathbf{e}_{i(t)}^\top \mathbf{w}_f) \label{eq:log_decay_model_rearranged}
\end{align}

Rearranging this equation, we define the target variable $y'$ and feature vector $\mathbf{x}'$ for a linear regression of the form $y' = \mathbf{x}'^\top \mathbf{w}_f$:

\begin{align} \label{eq:y_prime}
    y' =& \log(\tilde{\mu}_{i(t)} + \epsilon_{\text{log}}) - \log(r_t + \epsilon_{\text{log}}) \nonumber \\
    \mathbf{x}' =& N_{i(t)} \cdot \mathbf{e}_{i(t)}
\end{align}

Here, for numerical stability in computation, we introduce a small positive constant $\epsilon_{\text{log}} \ll r_t, \tilde{\mu}_{i(t)}$, preventing issues with $\log(0)$.

The online update of $\mathbf{w}_f$ proceeds as follows. First, a crucial condition is checked in Line 15: `If $r_t > 0$ and $\tilde{\mu}_{i(t)} > 0$ and $r_t \neq \tilde{\mu}_{i(t)}$ then`. This ensures that logarithmic transformation is valid and that actual decay has occurred, making the update meaningful.

If the conditions are met, $y'$ and $\mathbf{x}'$ are computed using the formulas in Equations \eqref{eq:y_prime}. Then we update the global decay parameter $\mathbf{w}_f$ through a standard online ridge regression framework (Line 16). After the loop of EM algorithm, sufficient statistics $\mathbf{A}$, $\mathbf{A}_f$ $\mathbf{b}$ and $\mathbf{b}_f$ are updated (Lines 19-23), utilizing the closed-form solutions for online ridge regression \cite{linucb}.

\subsection*{Application to Progressive Content Refinement}

In the preceding sections, we introduced DR-LinUCB, a versatile bandit algorithm designed for dynamic environments where rewards decay, and both arm context and decay parameters are learned using an EM algorithm. Now, we will demonstrate how this general algorithmic framework can be specifically applied to enhance LLMs through progressive content refinement.

We represent an LLM task as a tuple $(p_{\text{init}}, \text{Evaluator}())$, where $p_{\text{init}}$ is the original prompt that defines the LLM's primary objective (e.g., "Summarize the following article"), and $\text{Evaluator}()$ is a function that quantifies the quality or utility of the LLM's output ($y_t$) in response to a given prompt.

To apply DR-LinUCB to LLM tasks, each refinement prompt is treated as an "arm" within its framework. The process initializes $\mathcal{A}_0$ with $p_{\text{init}}$. In each step, the \textit{ArmGenerator} function (Algorithm 2) expands the set of available arms.
\begin{algorithm}[H]
\caption{ArmGenerator for Content Refinement}
\begin{algorithmic}[1]
\Function{ArmGenerator}{$\mathcal{A}_{t-1}, h_{t-1}$}
    \If{$t > 1$} $h_{t-1} \gets h_{t-1} \cup \{y_{t-1}\}$ \Else \space \Return $\{\}$ \EndIf\
    \State $p_{\text{feedback}} \gets \text{LLM}(h_{t-1})$
    \State $p_{\text{new}} \gets \text{LLM}(p_{\text{init}}, p_{\text{feedback}}, h_{t-1})$
    \State \Return $\{p_{\text{new}}\}$
\EndFunction
\end{algorithmic}
\end{algorithm}
This function leverages the LLM to generate a feedback prompt $p_{\text{feedback}}$ from the history $h_{t-1}$, and then generate a new refinement prompt $p_{\text{new}}$ based on $p_{\text{init}}$, $p_{\text{feedback}}$, and $h_{t-1}$. This self-refinement process draws inspiration from \cite{self-refine}. 

Additionally, we introduce an \textit{EvoArmGenerator} that employs EvoPrompt \cite{evoprompt} as a variant of arm generator. This evolutionary algorithm-based strategy explores a broader, more diverse set of refinement prompts. While self-refinement generation can converge to local optima, evolutionary algorithms excel at systematic exploration of wider solution spaces through mutation and crossover. This diversity is crucial for ablation studies comparing DR-LinUCB's learning-based selection against simpler strategies like random selection, especially when a rich pool of diverse arms is needed.

In Algorithm 1, the step for "Observe reward $r_t$" works as follows: First, the chosen prompt, $i(t)$, is given to the LLM, which then generates an output, $y_t$. The reward $r_t$ is subsequently calculated by feeding this output $y_t$ into the task's predefined $\text{Evaluator}()$ function, resulting in $r_t \gets \text{Evaluator}(y_t)$. These steps enable the application of the DR-LinUCB framework to LLM tasks for progressive content refinement.

\subsection*{Prompt Embedding}

To leverage our refinement prompts as arms within the DR-LinUCB framework, we employ a prompt embedding strategy. We begin by utilizing a fine-tuned MINILM \cite{MINILM} model \footnote{\url{https://huggingface.co/sentence-transformers/all-MiniLM-L6-v2}} to transform prompts into a high-dimensional embedding space. Then, we embed 200 prompts from the `awesome-chatgpt-prompts` \footnote{\url{https://huggingface.co/datasets/fka/awesome-chatgpt-prompts}} dataset into this space. Following this, we apply Principal Component Analysis (PCA) \cite{pca} to these embeddings to reduce their dimensionality, obtaining a 5-dimensional principal component vector. In this research, we leverage this fine-tuned sentence transformer model and the trained PCA model to embed new prompts and compress their dimensions before use in our bandit algorithm.

\begin{table*}[h!]
    \centering
    \begin{tabular}{cccc} 
        \toprule 
        \multirow{2}{*}{} & \multirow{2}{*}{} & \multicolumn{2}{c}{\textbf{Model}} \\
        \cmidrule(lr){3-4} 
        \textbf{Task} & \textbf{Algorithm} & \textbf{ChatGPT3.5-turbo} & \textbf{ChatGPT4o} \\
        \midrule 


        \multirow{7}{*}{\textbf{GSM8K}} 
        & Single Call & $0.187 \pm 0.059$ & $0.537 \pm 0.055$ \\
        & Random Exploration & $0.310 \pm 0.035$ & $0.647 \pm 0.049$ \\
        & JointLinUCB & $0.323 \pm 0.053$ & $0.610 \pm 0.061$ \\
        & EvoLinUCB & $0.340 \pm 0.053$ & $0.630 \pm 0.087$ \\  
        & Self-Refine & $0.687 \pm 0.075$ & $\mathbf{0.913 \pm 0.021}$ \\
        & REx & $0.530 \pm 0.017$ & $0.880 \pm 0.010$ \\
        & DR-LinUCB(Ours) & $\mathbf{0.790 \pm 0.035}$ & $0.900 \pm 0.010$ \\
        
        \midrule 

        \multirow{7}{*}{\textbf{Sentiment Inverse}} 
        & Single Call & $0.839 \pm 0.015$ & $0.950 \pm 0.016$ \\
        & Random Exploration & $0.935 \pm 0.028$ & $0.980 \pm 0.008$ \\
        & JointLinUCB & $0.922 \pm 0.010$ & $0.973 \pm 0.013$ \\
        & EvoLinUCB & $0.944 \pm 0.014$ & $0.984 \pm 0.002$ \\
        & Self-Refine & $0.879 \pm 0.009$ & $0.989 \pm 0.004$ \\
        & REx & $0.931 \pm 0.005$ & $0.985 \pm 0.008$ \\
        & DR-LinUCB(Ours) & $\mathbf{0.952 \pm 0.009}$ & $\mathbf{1.000 \pm 0.000}$ \\
        \bottomrule 
    \end{tabular}
    \caption{Performance summary (mean $\pm$ standard deviation) for each task, algorithm, and model. Metrics are defined in Section 5.}
    \label{tab:average_scores_example}
\end{table*}

\section{Evaluation}
We study two different domains that each involve complex natural language reasoning and generation.

\begin{enumerate}
    \item \textbf{Math Reasoning}:
GSM8K (Grade School Math 8k) \cite{cobbe2021gsm8k} is a challenging dataset of elementary-level math word problems. The task is to read a natural-language problem description and generate a step-by-step solution, involving numerical calculations and logical reasoning, to find the final numerical answer. It assesses a model's multi-step mathematical reasoning and its ability to convert linguistic information into a solvable procedure.

As a metric, we report the average success rate. A "success" is defined as generating the exact numerical answer defined in the dataset.

    \item \textbf{Sentiment Reversal}:
Sentiment Reversal is a long-form text style transfer task \cite{self-refine,sentimentreversal_data}. Given a text passage with a specific sentiment (e.g., negative), the goal is to rewrite the entire passage to a target sentiment (e.g., very positive), not merely reversing it. This task evaluates a model's ability for fine-grained content editing, requiring not only a complete shift in emotional tone but also an adjustment of sentiment intensity. It necessitates understanding the original meaning and tone, and then systematically altering vocabulary, phrasing, and narrative to achieve the specified opposing sentiment and intensity while maintaining coherence.

As a metric, we report the average scores. We first calculate the sentiment using a ModernBert-based Multilingual Sentiment Classification Model \cite{opensourcesnetiment}. If the predicted sentiment matches the target sentiment, a score of 1.0 is assigned. Otherwise, we employ an LLM as a judge to calculate a score reflecting the alignment of the generated text's sentiment to the target. The LLM model version used for judging in each experiment matches the version employed within that experiment.

\end{enumerate}

We use these tasks to study following research questions:

\noindent \textbf{RQ1:}
Can iterative-refinement achieve higher performance than a single-call strategy? 

\noindent \textbf{RQ2:} 
Can \textit{reward decay} reduce over-exploitation and contribute to total performance? 

\noindent \textbf{RQ3:} 
Does \textit{reward decay} accelerate the convergence and performance improvement of iterative refinement?

\noindent \textbf{RQ4:} 
Which approach provides better performance: ArmGenerator or EvoArmGenerator?

\noindent \textbf{RQ5:} 
Does the combination of joint LinUCB and reward decay achieve superior overall performance?

To investigate these questions, we study a range of baselines:

\begin{itemize}

    \item \textbf{Single Call}: LLM output from initial prompt only. Serves as a performance lower bound.

    \item \textbf{Random Exploration}: Uses \textit{EvoArmGenerator} but selects prompts uniformly at random. Evaluates the benefit of learning over pure exploration.

    \item \textbf{JointLinUCB}: Uses \textit{EvoArmGenerator} and LinUCB for prompt selection, but assumes stationary rewards, lacking decay modeling. Isolates decay's contribution.

    \item \textbf{EvoLinUCB}: This is a variant of the DR-LinUCB method that uses \textit{EvoArmGenerator}. This approach offers diverse prompt generation, but unlike the \textit{ArmGenerator}, it does not use LLM feedback during generation.
    \item \textbf{Self-Refine} \cite{self-refine}: Iterative refinement where LLM generates feedback and prompts. The prompt selection is greedy, typically choosing the latest generated prompt without adaptive learning. This corresponds to repeated \textit{ArmGenerator} use without sophisticated selection.

    \item \textbf{REx} (REfine, Explore, Exploit) \cite{rex}: Uses Thompson Sampling for content refinement. The refinement prompt is fixed, unlike our dynamic prompt selection.
\end{itemize}

For each experiment, we evaluated the models on 100 distinct samples for each of three random seeds. The reported results are averaged over these three independent runs, effectively evaluating performance on a total of 300 samples for each task and algorithm. For each sample, algorithm has up to six time steps.

\subsection*{The Value of Iterative LLM Interaction}
The comparison between Single Call and Random Exploration in Table \ref{tab:average_scores_example} demonstrates that multi-step iterative processes significantly enhance performance.

The Single Call baseline consistently yields the lowest scores. For instance, in GSM8K with ChatGPT3.5-turbo, Single Call scored 0.187. In contrast, Random Exploration shows a substantial improvement. For the same task, "Random Exploration" achieves 0.310, an approximate 65\% increase over Single Call. 

This clearly indicates that engaging the LLM in a multi-step iterative process leads to significant performance improvement over a single-shot approach. This observation answers RQ1, confirming that iterative LLM usage, even in its most basic form, yields higher performance than a single call strategy. It underscores the fundamental value of employing LLMs iteratively to refine and optimize their outputs.

\subsection* {DR-LinUCB Outperforms Baselines and SOTA}

The results in Table \ref{tab:average_scores_example} demonstrates that our proposed DR-LinUCB consistently achieves superior performance against both our ablation study baselines and state-of-the-art (SOTA) methods like Self-Refine and REx.

Specifically, with ChatGPT3.5-turbo, DR-LinUCB leads in both GSM8K ($0.790$) and Sentiment Inverse ($0.952$), significantly surpassing Self-Refine ($0.687$ for GSM8K, $0.879$ for Sentiment Inverse) and REx ($0.530 $ for GSM8K, $0.931$ for Sentiment Inverse). This highlights DR-LinUCB's capability in enhancing less powerful LLMs. With ChatGPT4o, DR-LinUCB achieves $0.900$ for GSM8K, outperforming REx ($0.880$) and closely trailing Self-Refine ($0.913$). In particular, for Sentiment Inverse with ChatGPT4o, DR-LinUCB achieves a perfect score of $1.000$, outperforming all other methods. We further discuss the feasibility and implications of this perfect score in Section 6.

These results unequivocally answer RQ5. The integration of these mechanisms allows DR-LinUCB to effectively learn optimal prompt arms while preventing over-exploitation, leading to robust and often best-in-class performance.

\begin{figure}[htb] 
    \centering 
     \hspace*{-0.8cm}
    \includegraphics[width=0.53\textwidth]{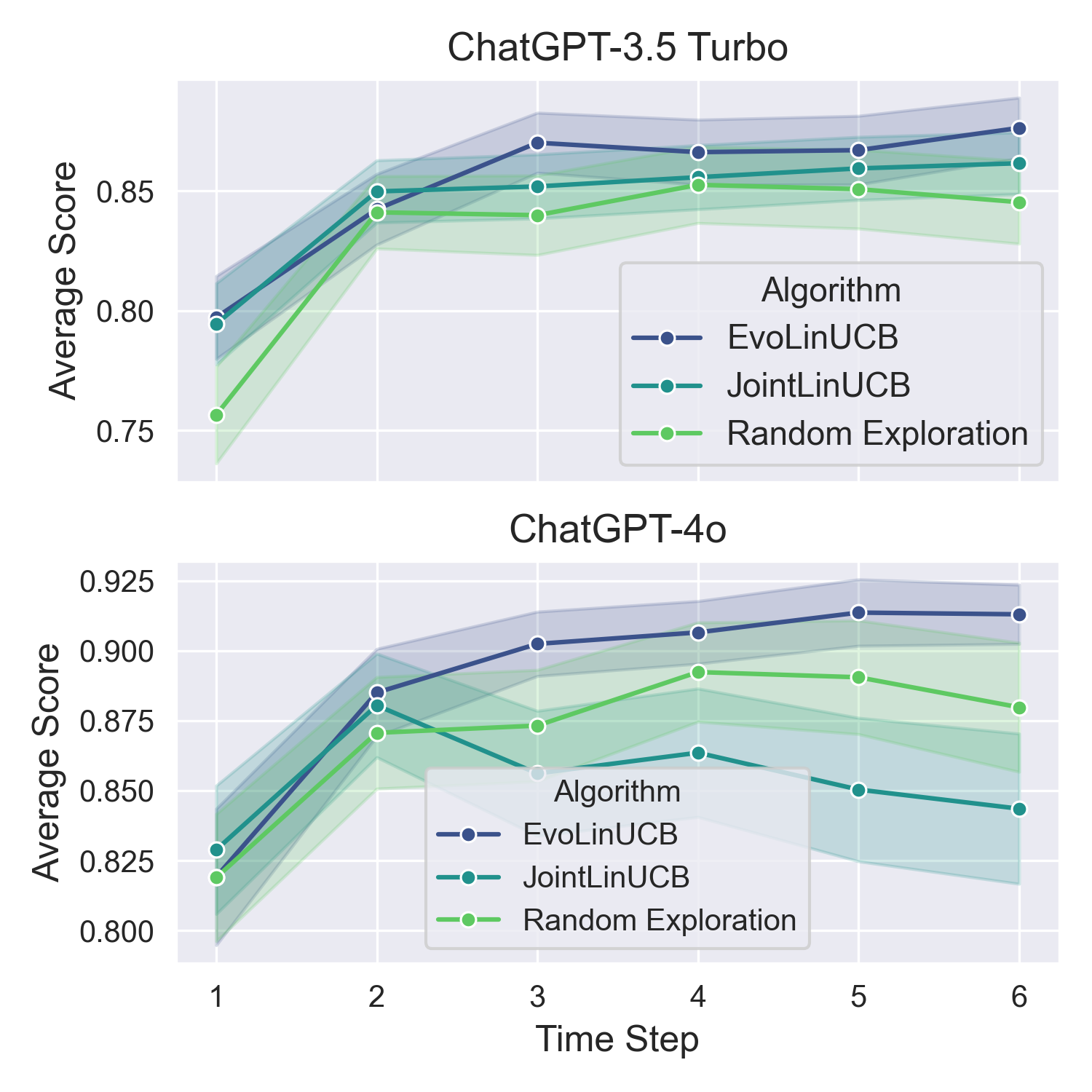}
    
    \caption{History of average scores in the Sentiment Reversal Task for ChatGPT-3.5 Turbo (top) and ChatGPT-4o (bottom). EvoLinUCB incorporates both arm decay and arm learning, while JointLinUCB only features arm learning. Random Exploration uniformly selects arms generated by EvoArmGenerator. Only instances not correctly answered in the first attempt are included, allowing for a comparison of iterative improvement effects. The shaded area indicates the standard error from three experimental runs.}
    \label{fig:score_history_combined} 
\end{figure}

\subsection*{Addressing RQ2 and RQ3: Insights from Algorithmic Component Analysis}
To ensure consistent conditions for prompt generation, our comparison focuses on three models that utilize the EvoPrompt Generator. Furthermore, only instances answered incorrectly in the first attempt are included to allow a comparison of iterative improvement effects.

As depicted in Figure \ref{fig:score_history_combined}, all three algorithms show comparable improvements up to the second iteration. However, from the third iteration onwards, EvoLinUCB consistently outperforms the others for both ChatGPT-3.5 Turbo and ChatGPT-4o. This superior performance of EvoLinUCB can be attributed to its incorporation of both reward decay and arm learning.

In particular, JointLinUCB performs worse than Random Exploration in the case of ChatGPT-4o. This observation supports our hypothesis that mechanisms preventing over-exploitation, like EvoLinUCB's decay or Random Exploration, improve effectiveness over solely exploiting estimated "good" arms. This finding directly addresses RQ2, indicating that reward decay effectively mitigates over-exploitation, leading to enhanced performance.

Moreover, by observing the score history, we can see that the scores converge relatively quickly. This rapid convergence provides an answer to RQ3, suggesting that effective improvements can be achieved within a limited number of iterative steps.

\subsection*{Suitability of Arm Generation Approaches}
DR-LinUCB (with \textit{ArmGenerator}) significantly outperforms EvoLinUCB in Sentiment Inverse ($0.952$ vs. $0.944$ with GPT-3.5) and achieves a substantial gain in GSM8K over JointLinUCB ($0.790$ vs. $0.323$). These results suggest that feedback-driven prompt exploration is generally more effective than evolutionary modification. 

However, this effectiveness is task-dependent; while Self-Refine excels in GSM8K with GPT-4o ($0.913$), it underperforms JointLinUCB in Sentiment Inverse, likely because iterative correction can introduce noise or suboptimal steering in simpler tasks. We conclude that \textit{ArmGenerator} is most robust when integrated with exploration-exploitation strategies.

\section{Conclusion}

This paper introduces Decaying Reward Joint LinUCB (DR-LinUCB), a novel bandit algorithm designed for iterative LLM content refinement. DR-LinUCB explicitly models the "saturation effect" through reward decay and jointly learns contextual and decay parameters. Our experiments on GSM8K and Sentiment Reversal tasks demonstrate DR-LinUCB's superior performance compared to various baselines and state-of-the-art methods. We confirmed that incorporating reward decay effectively prevents over-exploitation, leading to enhanced performance and accelerated convergence.

DR-LinUCB offers a significant advancement in optimizing LLM iterative refinement, providing a principled approach to balance exploration and exploitation in dynamic environments where the utility of refinement strategies can diminish over time.

\newpage

\section{Limitations}

While our proposed DR-LinUCB framework shows promising results, several limitations remain to be addressed in future work.

\paragraph{Feasibility of Score 1.0 in Sentiment Reversal Task}
Historically, LLMs have struggled to achieve 100\% accuracy on simple tasks, largely due to inherent flaws in existing benchmarks. The Platinum benchmark \cite{vendrow2025largelanguagemodelbenchmarks} demonstrates that many benchmarks contain significant errors; for instance, in SVAMP \cite{svamp}, a question was mislabeled with an incorrect solution. In contrast, our sentiment reversal task evaluates performance using a pretrained sentiment analysis model and an LLM judge, effectively eliminating the issue of mislabeled ground truth. 

While the task is relatively simple—with even the Self-Refine model achieving a high score of 0.989—the fact that 100\% accuracy is achievable aligns with the findings of the Platinum Benchmarks paper regarding carefully curated tasks. However, it remains a limitation that our current evaluation is focused on this feasible task, and further validation on more complex benchmarks where 100\% accuracy is not yet reachable is required.

\paragraph{Minimizing Cost of LLM Interactions}
The iterative nature of our approach, combined with multiple LLM calls for ArmGenerator and reward evaluation within each timestep, can lead to significant computational overhead and increased API costs. In real-world, high-throughput scenarios, the latency and expenses associated with these repeated inferences could be limiting factors. While our reward decay model implicitly helps minimize unnecessary calls by discouraging the over-exploitation of ineffective prompts, the framework is not yet explicitly optimized for this purpose. Future research should investigate how to leverage decay modeling more strategically to reduce the total number of LLM interactions required, improving overall cost-effectiveness.

\paragraph{Sensitivity to Hyperparameters}
Like many bandit algorithms, DR-LinUCB relies on hyperparameters such as $C$, $\lambda$, $\lambda_f$, $\epsilon_{\text{log}}$, and $\epsilon_{\text{decay}}$. The performance of the algorithm can be sensitive to the tuning of these parameters. While we have identified effective settings for our specific tasks, optimal tuning for diverse LLM applications might require extensive experimentation. The development of adaptive or self-tuning mechanisms for these hyperparameters remains an important direction to enhance the robustness and ease of use of DR-LinUCB.


\bibstyle{acl_natbib} 
\bibliography{custom}          

\section{Appendix} 
\label{sec:appendix} 

\subsection{Implementation Details}
\label{subsec:implementation_details} 
This appendix provides a comprehensive overview of the experimental setup and specific hyperparameter configurations used in our study to ensure reproducibility. 

\subsubsection{Hardware and Software Environment}
All experiments were conducted on a MacBook Pro (13-inch, 2020) equipped with a 1.7 GHz Quad-Core Intel Core i7 processor, Intel Iris Plus Graphics 645 (1536 MB), and 16 GB of 2133 MHz LPDDR3 memory. For the LLM interactions, we used OpenAI's API and client library.
\subsubsection{Hyperparameters}
The specific hyperparameters used for each algorithm are detailed below. 

\begin{itemize}
    \item \textbf{DR-LinUCB:} For the DR-LinUCB algorithm, we set the maximum number of iterations to \texttt{max\_itr=5}, and the early stopping score to \texttt{early\_stop\_score=1.0}.The number of iterations for parameter learning was \texttt{T\_em=2}. The dimension of the context vector was \texttt{d\_dim=5}, and the constant for the confidence bound was \texttt{C\_constant=0.5}.   Regularization parameters were set to \texttt{lambda\_theta=0.1} and \texttt{lambda\_wf=0.1}. Small epsilon values for logarithmic and decay calculations were \texttt{epsilon\_log\_val=$1 \times 10^{-6}$} and \texttt{epsilon\_decay\_val=$1 \times 10^{-6}$} respectively.

    \item \textbf{Rex:} The Rex algorithm utilized a confidence parameter \texttt{C=20.0}, a maximum of \texttt{max\_llm\_calls=5} to the LLM, and an early stopping score of \texttt{early\_stop\_score=1.0}.

    \item \textbf{Self-Refine:} For the Self-Refine process, we configured \texttt{max\_iterations=5} and an \texttt{early\_stop\_score=1}.
\end{itemize}

\subsection{Prompts}

\subsubsection{GSM8K}

For the GSM8K (Grade School Math 8K) task, we employ a similar multi-stage prompting framework designed to enhance the accuracy and robustness of problem-solving. This framework consists of an initial generation prompt, a feedback generation prompt, and a refinement prompt, iteratively improving the mathematical reasoning.

\subsubsection{Initial Problem Solving Prompt}
The first stage aims to generate an initial solution to the mathematical word problem. The prompt guides the LLM to provide a detailed Chain-of-Thought (CoT) before stating the final answer:
\begin{lstlisting}[caption=Initial Problem Solving Prompt]
Solve the following mathematical word problem. Output the final answer only after showing your detailed step-by-step reasoning (Chain-of-Thought).

Question: {question}

Answer:
\end{lstlisting}
This prompt instructs the LLM to solve the given `question` by first producing a step-by-step reasoning process, followed by the final numerical answer.

\subsubsection{Feedback Generation Prompt}
After the initial solution attempt, a critical feedback stage is introduced. This prompt directs the LLM to act as a reviewer, identifying potential issues in the generated reasoning without revealing the correct answer:
\begin{lstlisting}[caption=Feedback Generation Prompt]
Review the given mathematical word problem and the current reasoning/answer. Provide detailed, step-by-step feedback focused on identifying potential calculation errors, logical flaws, or misinterpretations of the question. Do not state the final answer. Current reasoning/answer:
\end{lstlisting}
The LLM is tasked with providing constructive feedback on the `current reasoning/answer`, specifically looking for `calculation errors`, `logical flaws`, or `misinterpretations` of the problem. A crucial directive is to `not state the final answer`, ensuring the feedback mechanism does not directly provide the solution.

{Refinement Prompt}
The refinement stage utilizes the generated feedback and the history of previous attempts to iteratively improve the solution. This prompt is designed to guide the LLM towards a correct and well-structured answer:
\begin{lstlisting}[caption=Refinement Prompt]
**[Refinement Task: Math Word Problem (GSM8K)]**

**Question:** {question}

Review the provided question, the previous attempts at solving it, and the detailed feedback history. Your task is to **refine the entire reasoning chain and the final answer**.

**Refinement Directives:**
* **Calculation Accuracy:** Strictly re-check every arithmetic step. Identify and correct any calculation errors (miscounts, incorrect multiplication/division, etc.).
* **Logical Consistency:** Verify the logical flow of the Chain-of-Thought (CoT). Ensure each step is derived correctly from the previous one and aligns with the question's premise.
* **Answer Format:** After the refined step-by-step reasoning, output the final numerical answer clearly. The final output must conclude with the phrase: **'The final answer is [NUMBER]'**.

**History of Attempts and Feedback:**
---
{history}
---

**Refined Reasoning and Final Answer:**
\end{lstlisting}
This comprehensive refinement prompt explicitly defines the task as `[Refinement Task: Math Word Problem (GSM8K)]` and provides specific `Refinement Directives`. These directives guide the LLM to focus on `Calculation Accuracy`, `Logical Consistency` of the Chain-of-Thought (CoT), and `Answer Format`. It emphasizes the need to re-check all arithmetic steps and ensure the logical flow aligns with the `question`'s premise. Crucially, it mandates a specific output format for the final answer: \texttt{'The final answer is [NUMBER]'}. The prompt also incorporates a `History of Attempts and Feedback`, allowing the LLM to learn from past errors and improve its reasoning iteratively.

\subsubsection{Sentiment Reversal}

Our approach to sentiment reversal leverages a multi-stage prompting strategy with a Large Language Model (LLM). This strategy involves an initial sentiment inversion prompt, a feedback generation prompt, and a refinement prompt, designed to iteratively improve the quality and accuracy of the inverted review.

\subsubsection{Initial Sentiment Inversion Prompt}
The first stage involves generating an initial inverted review. The prompt is structured to clearly instruct the LLM on the primary task:
\begin{lstlisting}[caption=Initial Sentiment Inversion Prompt]
Invert the sentiment of the following review.

Original Review: {original_review_text}

Inverted Review:
\end{lstlisting}
This prompt directly asks the LLM to perform the sentiment reversal on the provided \texttt{original\_review\_text} and to output the \texttt{Inverted Review}.

\subsubsection{Feedback Generation Prompt}
Following the initial inversion, a feedback mechanism is employed to evaluate the generated inverted review. This feedback is crucial for guiding subsequent refinements. The prompt for generating feedback is as follows:
\begin{lstlisting}[caption=Feedback Generation Prompt]
Review the given original review ({original_review_text}...) and the current inverted review.
Provide concrete suggestions for improvement, focusing on whether the sentiment is properly
inverted and if the expression is natural.
Consider the Original Sentiment: {original_sentiment} and the Target Sentiment: {target\_sentiment}.
Current inverted review:
\end{lstlisting}
This prompt instructs the LLM to act as a critic, providing constructive feedback. It explicitly asks for suggestions focusing on sentiment inversion accuracy and naturalness of expression, taking into account the \texttt{original\_sentiment} and the desired \texttt{target\_sentiment}.

\subsubsection{Refinement Prompt}
The final stage utilizes the feedback to refine the inverted review. This iterative refinement process aims to converge on a high-quality, sentiment-inverted text. The refinement prompt is designed to guide the LLM through this process:
\begin{lstlisting}[caption=Refinement Prompt]
Please return only inverted reviews and don't return suggestions.
For the task of inverting the sentiment of the Original Review: {original\_review\_text}...,
refine the review based on the history of previous inverted reviews and feedback.
Consider all provided feedback and ensure the refined text has a {target\_sentiment} sentiment.
Current sentiment is {self.sentiment}.
if sentiment is not enough (e.g., target is very positive is current is positive),
change sentiment polarity and change Intensifiers / Amplifiers
Please consider current history:
\end{lstlisting}
This prompt emphasizes that the LLM should only output the refined inverted review, without additional suggestions. It directs the LLM to leverage the \texttt{history of previous inverted reviews and feedback} to ensure the refined text aligns with the \texttt{target\_sentiment}. It also provides explicit instructions for cases where the current sentiment is not sufficiently strong, suggesting the modification of sentiment polarity and the use of intensifiers/amplifiers, thereby promoting more robust sentiment manipulation. The prompt also implicitly references an internal \texttt{self.sentiment} variable and encourages consideration of the \texttt{current history} of refinements.

\subsubsection{Prompts for generating arms}

\subsubsection{\texttt{EvoArmGenerator}}
This meta-prompt guides the Large Language Model (LLM) to perform evolutionary operations (mutation and crossover) on a target prompt. It leverages information from three "donor" prompts and an experimental history to create a refined prompt. The process is designed to optimize existing prompts while preserving their core objective.
\begin{lstlisting}[caption=EvoArmGenerator]
You are an expert prompt optimizer. Your task is to perform an evolutionary operation
on the target prompt using three donor prompts.
1.Identify the key elements and style differences between Donor 1 and Donor 2.
Donor 1 (Pr1): {donor1_prompt}
Donor 2 (Pr2): {donor2_prompt}
2.Mutate the different parts, considering the history of experiment: {history}
3. Combine the different parts with Prompt 3, selectively replace it with the different parts in Step 2 and generate a new prompt. Don't use Donar 1 and Doner 2 as pronoun
4. Cross over the prompt in the Step 3 with the following basic prompt and generate a final prompt bracketed with <prompt> and </prompt>
5. Do not break original goal (if it's mathmatical calculation, don't mutate number and formula). Mutate expression and approach to accomlish original goal.
6. When mutating, simplify prompt as well (especially for mathmatical calculation)
\end{lstlisting}

The LLM, acting as an "expert prompt optimizer," is instructed to analyze the stylistic and elemental differences between `donor1\_prompt` and `donor2\_prompt`. It then mutates relevant parts based on the `history` of experiments, combines these with an implicit "Prompt 3," and performs a crossover operation with a "basic prompt." Critical directives include maintaining the `original goal` (e.g., preserving numerical values and formulas in mathematical calculations) while focusing on mutating `expression and approach`. Additionally, simplification of the prompt during mutation is encouraged, particularly for mathematical tasks. The final output prompt is required to be enclosed within `<prompt>` and `</prompt>` tags.

\subsubsection{History-Based Novel Prompt Generator (\texttt{ArmGenerator})}
This meta-prompt aims to generate entirely new prompts that fundamentally deviate from previous solutions, ensuring novelty while strictly adhering to the original task's objective. It leverages the full `Experiment History` to strategically guide the LLM towards innovative approaches.
\begin{lstlisting}[caption=ArmGenerator]
You are a seasoned strategist overseeing prompt evolution. For the given "original prompt," leverage the following historical information to its fullest extent and generate a **novel prompt that distinctly deviates from previous approaches**.

Experiment History: {history}

Instructions for generating the new prompt:
1.  **Fundamental Shift in Approach**: Deeply analyze the history above and devise a completely new approach that intentionally departs from the expressions, structures, or thought patterns adopted in previous attempts.
2.  **Absolute Maintenance of Goal**: The ultimate goal defined by the original prompt, including specific numbers, formulas, or data in mathematical calculations, must not be altered for any reason. These must be entirely preserved in the new prompt.
3.  **Refinement and Simplification of Expression**: The new approach should be more refined and concise to achieve the original goal. Especially in cases involving complex calculations or logic, eliminate redundancy and pursue the most efficient and clear expression possible.

4. Generate a final prompt bracketed with <prompt> and </prompt>
\end{lstlisting}

Acting as a "seasoned strategist," the LLM is tasked with generating a `novel prompt` that `distinctly deviates from previous approaches` by thoroughly analyzing the `Experiment History`. The core instructions emphasize a `Fundamental Shift in Approach`, meaning a deliberate departure from past expressions, structures, or thought patterns. Despite this push for novelty, the `Absolute Maintenance of Goal` is paramount, ensuring that critical elements like numbers and formulas in mathematical contexts remain unchanged. Furthermore, the new approach should exhibit `Refinement and Simplification of Expression`, eliminating redundancy and promoting efficiency, especially for complex tasks. As with the evolutionary prompt, the final output must be enclosed within $<prompt>$ and $</prompt>$ tags.

\end{document}